\documentclass[final]{cvpr}

\usepackage{times}
\usepackage{epsfig}
\usepackage{graphicx}
\usepackage{amsmath}
\usepackage{amssymb}
\usepackage{placeins}

\usepackage{mmstyle}

\usepackage{multirow}

\usepackage[pagebackref=true,breaklinks=true,colorlinks,bookmarks=false]{hyperref}

\def\cvprPaperID{3516} 
\def\confYear{CVPR 2021}

\begin{document}

\title{Unsupervised 3D Shape Completion through GAN Inversion}

\author{Junzhe Zhang$^{1,3}$ \quad
Xinyi Chen$^{2,3}$ \quad
Zhongang Cai$^{3,4}$ \quad
Liang Pan$^{1}$ \quad
\\Haiyu Zhao$^{3,4}$ \quad
Shuai Yi$^{3,4}$ \quad
Chai Kiat Yeo$^{2}$ \quad
Bo Dai$^{1}$ \quad
Chen Change Loy$^{1}$ \\
$^{1}$S-Lab, Nanyang Technological University  \hspace{8pt} $^{2}$Nanyang Technological University  \hspace{8pt} \\ $^{3}$SenseTime Research \hspace{8pt} $^{4}$Shanghai AI Laboratory  \\

{\tt\small {\{junzhe001,xchen032\}@e.ntu.edu.sg,\{liang.pan,asckyeo,bo.dai,ccloy\}@ntu.edu.sg}}\\
{\tt\small {\{caizhongang,zhaohaiyu,yishuai\}@sensetime.com}} \\

\normalsize
\textcolor{magenta}{\url{https://junzhezhang.github.io/projects/ShapeInversion/}}
\vspace{-5mm}
}

\maketitle


\begin{abstract}
Most 3D shape completion approaches rely heavily on partial-complete shape pairs and learn in a fully supervised manner. Despite their impressive performances on in-domain data, when generalizing to partial shapes in other forms or real-world partial scans, they often obtain unsatisfactory results due to domain gaps. In contrast to previous fully supervised approaches, in this paper we present ShapeInversion, which introduces Generative Adversarial Network (GAN) inversion to shape completion for the first time. ShapeInversion uses a GAN pre-trained on complete shapes by searching for a latent code that gives a complete shape that best reconstructs the given partial input. In this way, ShapeInversion no longer needs paired training data, and is capable of incorporating the rich prior captured in a well-trained generative model. On the ShapeNet benchmark, the proposed ShapeInversion outperforms the SOTA unsupervised method, and is comparable with supervised methods that are learned using paired data. It also demonstrates remarkable generalization ability, giving robust results for real-world scans and partial inputs of various forms and incompleteness levels. Importantly, ShapeInversion naturally enables a series of additional abilities thanks to the involvement of a pre-trained GAN, such as producing multiple valid complete shapes for an ambiguous partial input, as well as shape manipulation and interpolation.

\end{abstract}


\section{Introduction}
\label{sec:intro}

\begin{figure*}[h!]
\vspace{-2mm}
\begin{center}
  \includegraphics[width=0.99\linewidth]{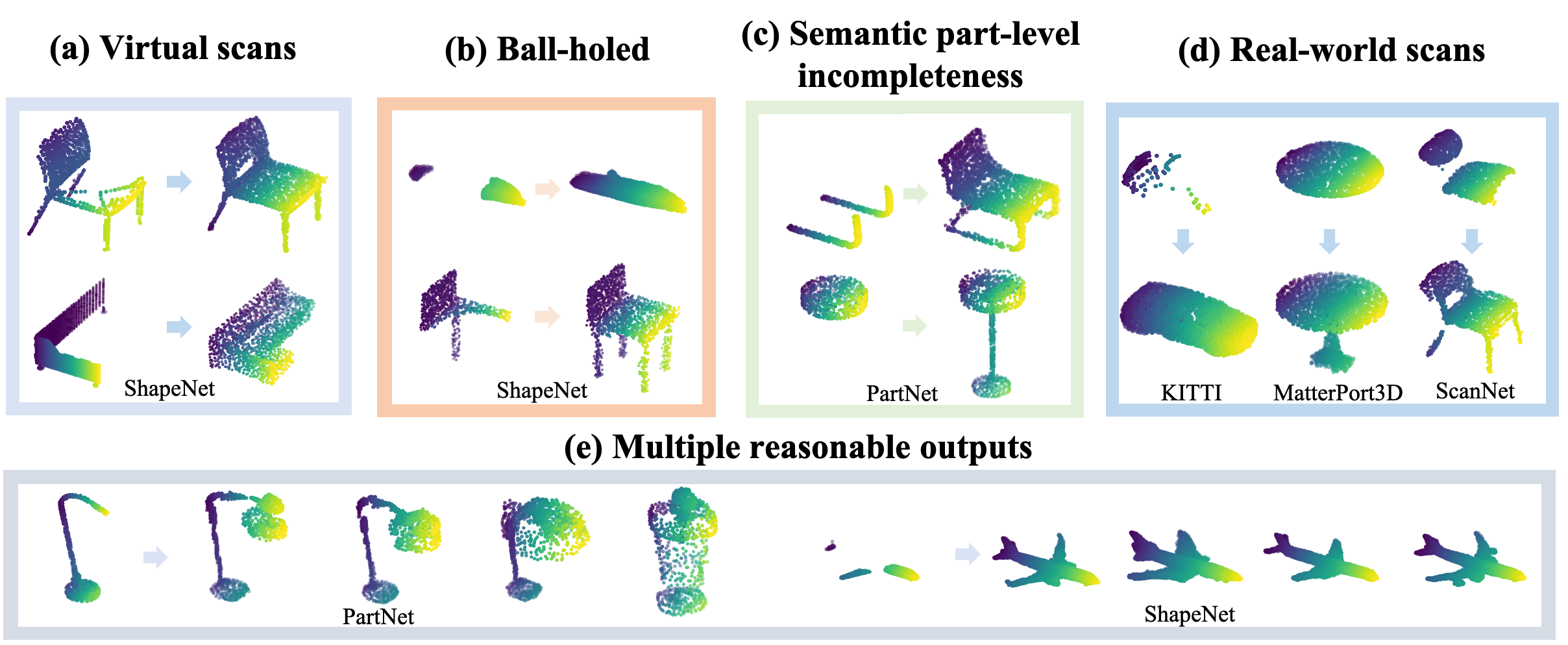}
\end{center}
\vspace{-4mm}
  \caption{ShapeInversion incorporates the prior captured by a well-trained GAN. It shows exceptional generalization ability for shape completion: is invariant to partial form changes, \ie, \textbf{(a)} virtual scans, generated by back-projecting 2.5D depth images into 3D, \textbf{(b)} ball-holed partial shapes, generated by removing points within a random ball from a complete shape (PF-Net~\cite{huang2020pf}), \textbf{(c)} semantic part-level incompleteness, generated by randomly removing some semantic parts (PartNet~\cite{mo2019partnet}); and generalizes well to \textbf{(d)} real-world scans. Moreover, it can give \textbf{(e)} multiple valid outputs when there is ambiguity in the partial shape}
\vspace{-3mm}
\label{fig:teaser}
\end{figure*}

3D shape completion estimates the complete geometry from a partial shape in the form of a partial point cloud,
and is important to many downstream applications such as robotics navigation~\cite{engel2014lsd, mur2015orb} and scene understanding~\cite{dai2018scancomplete, hou20193d}.
Most works~\cite{tchapmi2019topnet, liu2019morphing, huang2020pf,wang2020cascaded,wen2020point} for shape completion are trained in a fully supervised manner with paired partial-complete data.
While they obtain promising results on in-domain data,
it is challenging for these methods to generalize to out-of-domain data,
which are real-world scans or data with different partial forms,
as shown in Fig.~\ref{fig:teaser} (a)-(d).

We take an unsupervised approach in this study. Inspired by the success of GAN inversion in 2D tasks such as image restoration and editing,
we propose to apply GAN inversion to 3D shape completion for the first time,
which we refer to as \textbf{ShapeInversion}.
Specifically, given a partial input,
ShapeInversion looks for a latent code in the GAN's latent space that gives a complete shape that best reconstructs the input.
By incorporating prior knowledge stored in the pre-trained GAN,
no assumptions on the input partial forms are made,
thus ShapeInversion generalizes well to inputs of various partial forms and real-world scans.
Moreover, the involvement of GAN in ShapeInversion brings several side-benefits,
including giving multiple reasonable complete shapes for some partial input,
as well as shape jittering and shape manipulation.

While ShapeInversion shares some similarity with GAN inversion methods for 2D images,
the former possesses several intrinsic challenges due to the nature of 3D data:
(1) Unlike 2D images that follow a grid-like structure,
where the positions of pixels are well defined,
point clouds of different 3D shapes are highly unstructured.
Often,
GANs trained on 3D shapes would generate point clouds with significant non-uniformity,
\ie, points are unevenly distributed over the shape surface.
Such non-uniformity may lead to shapes with undesired holes, undermining the completeness of our predictions.
(2) The unordered nature of point clouds makes the completion task significantly different from 2D image inpainting.
In 2D image inpainting, one can easily measure the reconstruction consistency between the visible regions of partial input and predicted output given the lattice-aligned pixel correspondences. 
Such comparison is challenging in 3D shape completion since the corresponding regions of two 3D shapes may reside at different locations in the 3D space. Without accurate point correspondences, GAN inversion would suffer from poor reconstruction and in turn jeopardize the shape completion task.

We present two new components to address these unique challenges.
First,
to improve the uniformity of estimated point clouds,
we introduce a simple and effective uniform loss, \textbf{PatchVariance}. 
The loss samples small patches, to ensure the planar assumption, over the object surface, 
and penalize the variance of average distances between the patch centers and their respective nearest neighbors.
Unlike existing methods~\cite{li2019pu,yu2018pu} that are typically conducted at the patch level,
ours is a soft regularizer that enhances uniformity at the object level on-the-fly while training GANs.
As a result, we achieve improved uniformity across all categories,
ranging from bulk to fine structures while preserving the shape plausibility and variety.

Second, we devise an effective masking mechanism, \textbf{k-Mask}, to estimate the point correspondences between the partial input and predicted shape.
To mitigate the ambiguous correspondences caused by the unordered nature of point cloud, our method lets each point in the partial input look for its k-nearest neighbors from the predicted shape. The indices of all these k-nearest neighbors define the mask of the visible regions, from which we can compute for reconstruction loss.
Our method is dynamic, thus performing better than baseline approaches that use predefined voxels or distance thresholds. It shows high robustness even when the semantics parts between the partial input and the predicted shape are not within a close vicinity in the space. 

ShapeInversion demonstrates compelling performance for shape completion in different scenarios.
First, on a common benchmark derived from ShapeNet,
it outperforms the SOTA unsupervised method pcl2pcl~\cite{chen2019unpaired} by a significant margin,
and is comparable to various supervised methods.
Second, our method shows considerable generalization ability and robustness
when it comes to real-world scans or variation in partial forms and incompleteness levels,
whereas supervised methods exhibit significant performance drops due to domain mismatches.
Third, given more extreme incompleteness that causes ambiguity,
our method is able to provide multiple valid complete shapes,
all of which remain faithful to the visible parts presented in the partial input.
%


\section{Related Work}

\noindent
\textbf{3D Shape Completion}.
3D shape completion has played an important role for robotics ~\cite{engel2014lsd,mur2015orb} and perception~\cite{dai2018scancomplete,hou20193d}.
Since the pioneering work PCN~\cite{yuan2018pcn}, point cloud-based shape completion has seen significant development compared to other representation forms like meshes and voxel grids, due to its flexibility and popularity as a raw data format.
Most existing approaches are trained in a fully-supervised manner with partial shapes of a particular form~\cite{dai2017shape,huang2020pf,chen2019unpaired,mo2019partnet,wen2020point,zhang2020detail}, and paired complete shapes. Owing to the coarse-to-fine strategy~\cite{tchapmi2019topnet, liu2019morphing, huang2020pf,wang2020cascaded,xie2020grnet}, they achieve impressive results on in-domain data, but may fail to sufficiently generalize to real-world scans or partial shapes in other forms.
Recently, pcl2pcl~\cite{chen2019unpaired} proposes an unsupervised method with unpaired data, \eg, complete shapes obtained from 3D models and real-world partial scans. It trains two separate auto-encoders, for reconstructing complete shapes and partial ones respectively, and learns a mapping from the latent space of partial shapes to that of the complete ones. 
In view of ambiguity at high incompleteness levels, its follow-up work~\cite{wu2020multimodal} is able to output multiple plausible complete shapes, conditioned by an additional latent vector drawn from Gaussian distribution.
Our approach also lies in the unsupervised regime, and can also give multiple reasonable complete shapes thanks to the involvement of a pre-trained GAN. Moreover, we achieve more faithful results, particularly for real scans.

\noindent
\textbf{GAN Inversion}.
State-of-the-art GANs, \eg, BigGAN~\cite{brock2018large} and StyleGAN~\cite{karras2019style}, are typically trained on a large number of images and capture rich knowledge of images including low-level statistics, image semantics, and high-level concepts. GAN inversion uses a well-trained GAN as effective prior to reconstruct images with high-fidelity. This appealing nature of GAN prior has been extensively exploited on various image restoration and manipulation tasks \cite{bau2019seeing,bau2020semantic,pan2020exploiting,gu2020image}.
In general, the method aims to find a latent vector that best reconstructs the given image with a pre-trained GAN.
Typically, the latent vector can be optimized based on gradient descent~\cite{ma2018invertibility,lipton2017precise}, or projected by an extra encoder from the image space \cite{zhu2016generative,lei2019inverting}. Moreover, the introduced encoder can serve as a better initialization prior to gradient descent ~\cite{bau2020semantic}. Zhu \etal \cite{zhu2020domain} learn a domain-guided encoder, which is used to regularize the latent vector optimization for semantically meaningful editing. 
While mainstream approaches fix the parameters of the generator during inversion, recent approaches chose to perturb~\cite{bau2019seeing} or fine-tune~\cite{pan2020exploiting} the generator when updating the latent vector to address the gap between the approximated manifold and the real one. 
Our approach is the first to apply GAN inversion to shape completion. Unlike image-based tasks where the degradation transform is typically straightforward, transforming a complete shape into a partial one in the 3D space is ill-posed.


\section{Method}

A GAN that is well-trained on 3D shapes of a particular category, \eg, chairs or cars, captures rich shape geometries and semantics of this distribution.
In this study, we wish to incorporate a well-trained GAN as an effective prior for shape completion, in particular, to handle partial shapes of a wide range of varieties and to generalize to unseen shapes.
The GAN prior can be exploited through GAN inversion. Despite its notable success in various image restoration and manipulation tasks, it has not been explored for shape completion.

\begin{figure}[t]
\begin{center}
\vspace{-1mm}
 \includegraphics[width=0.99\linewidth]{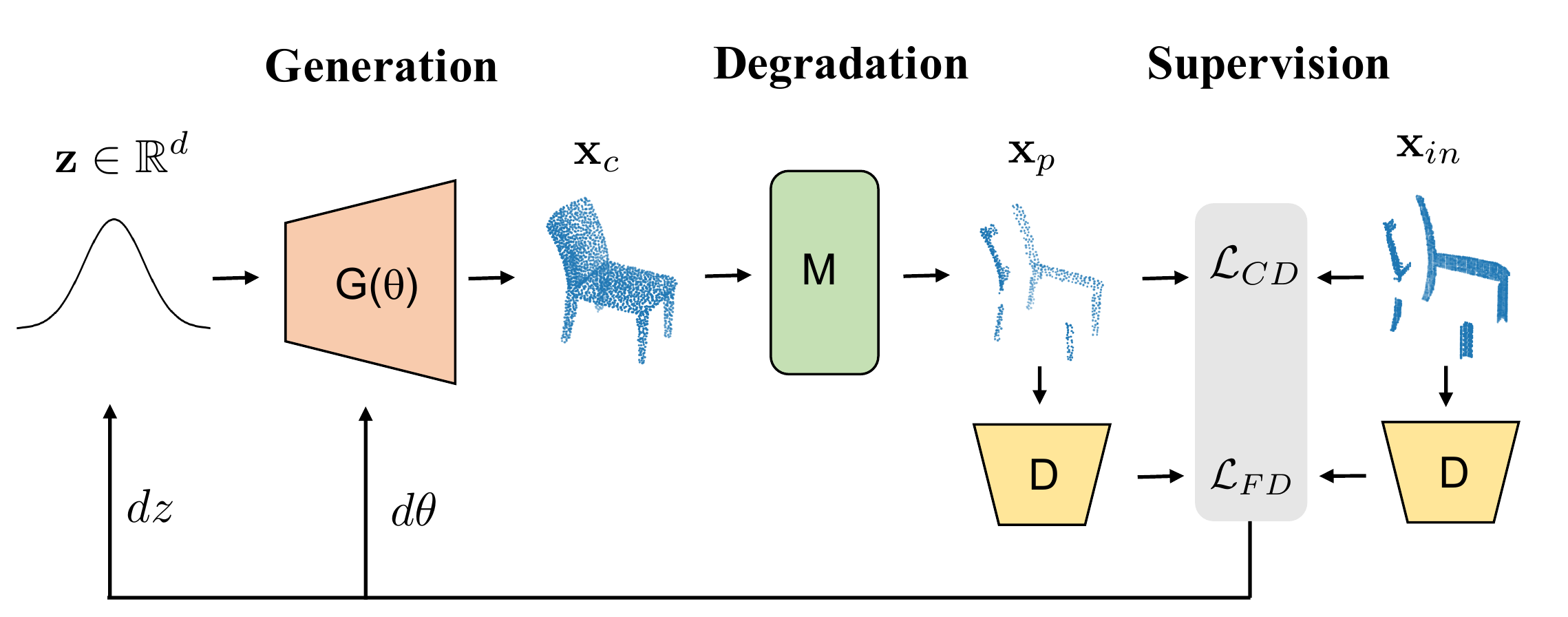}
\end{center}
\vspace{-3mm}
 \caption{GAN inversion for shape completion. A latent vector $\vz$ is used by the pre-trained generator $G$ to reconstruct a complete shape $\vx_c$. The degradation function $M$ (Sec.~\ref{sec:mask}) then transforms $\vx_c$ into a partial shape $\vx_p$. The supervision signal includes the Chamfer Distance and the Feature Distance (Sec.~\ref{sec:loss}) between $\vx_p$ and the input partial shape $\vx_{in}$. ShapeInversion looks for a latent vector $\vz$ and finetunes the parameters $\theta$ of $G$ that best reconstruct the complete shape corresponding to $\vx_{in}$ via gradient descent}
 \vspace{-3mm}
\label{fig:diagram}
\end{figure}

Here, we formally introduce the use of GAN inversion in our task. After a generator $G$ with parameters $\theta$ is trained on 3D shapes in the form of point clouds,
it can generate a shape $\vx_c \in \Rbb^{m\times3}$ from a latent vector $\vz \in \Rbb^d$. GAN inversion aims to find the latent vector that best reconstructs a given shape $\vx_{in}$ using $G$:
\begin{equation}
  \vz^{*} = \argmin_{\vz \in \Rbb^d} \cL(G(\vz; \vtheta),\vx_{in}), \quad
  \vx_{c}^{*} = G(\vz^{*}; \vtheta)
\end{equation}

While mainstream approaches usually fix the generator during inversion, we follow the very recent approaches~\cite{pan2020exploiting,bau2019seeing} to fine-tune the generator while updating the latent vector on-the-fly,
which is shown to improve the results of GAN inversion.
Thus, the formulation becomes:
\begin{equation}
  \vtheta^*, \vz^* = \argmin_{\vz, \vtheta} \cL(G(\vz;\vtheta),\vx_{in}) \label{eq:dgp}
\end{equation}

The inversion process starts with an initialization stage, in which hundreds of latent vectors are sampled randomly, and the $z$ with the smallest $\cL$ value is selected as the initial value for fine-tuning.
Then both $z$ and $\theta$ are updated via gradient descent according to Eq.~\eqref{eq:dgp}.
In the scenario of shape completion, we aim to infer a complete shape $\vx_{c}$ from a given partial shape $\vx_{in}$, where the distance is computed at the observation space, \ie, we would need to transform a complete shape into a partial form via a degradation function $M$, as shown in Eq.~\eqref{eq:shapeinversion}. Thus, it is essential for $M$ to provide precise point correspondences for the sake of an accurate reconstruction loss.
The inversion stage is shown in Fig.~\ref{fig:diagram}.
\begin{equation}
\label{eq:shapeinversion}
 \vz^{*} = \argmin_{\vz \in \Rbb^d} \cL(M(G(\vz; \vtheta)),\vx_{in})
\end{equation}

\begin{figure}[t]
\vspace{-2mm}
\begin{center}
 \includegraphics[width=0.99\linewidth]{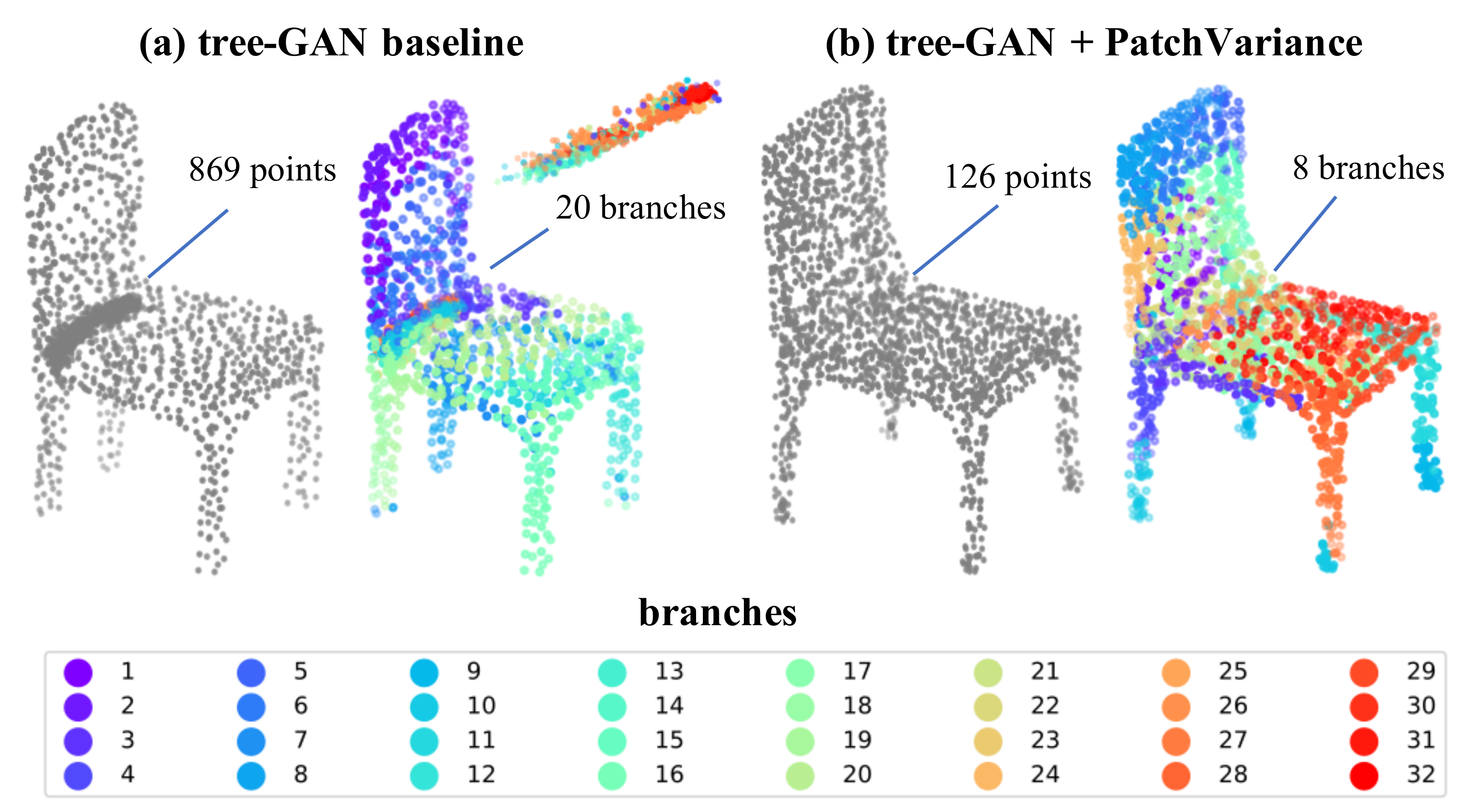}
\end{center}
\vspace{-4mm}
 \caption{Visualization of uniformity with the use of PatchVariance. Darker regions in the grayscale image have a higher density of points. tree-GAN uses a tree structure to generate 3D shapes. We group points into \textit{branches} by their parent nodes: branches that are more distantly related are further apart on the color spectrum. For the tree-GAN baseline, nearly half of the points and 20 out of the 32 branches cluster around the joint of different parts, with fewer branches to cover the rest of the shape surface}
 \vspace{-3mm}
\label{fig:uniform_loss}
\end{figure}

\subsection{Enhancing Point Cloud Uniformity}
\label{sec:uniform}

Compared to images where the generated pixels are arranged in a regular lattice, 3D shapes are represented by points in a continuous 3D space without a common structure. As a result, 3D GANs often generate point clouds with significant non-uniformity, where the points are often unevenly distributed over the shape surface.
Such non-uniformity is detrimental to shape completion given the number of points in each point cloud is fixed (typically 2048 for existing GANs): point concentration in one region inevitably leads to sparsity or even holes in other regions.

\noindent
\textbf{tree-GAN as a Case Study}. The latest state-of-the-art point cloud generation method, tree-GAN~\cite{shu20193d}, employs a tree-structured graph convolution network (TreeGCN) as the generator, where the information passes from the ancestor nodes instead of the neighbor nodes.
As branching occurs between every two layers in TreeGCN, the child nodes sharing the same parent node would be more geometrically related to each other.
Although it outperforms previous approaches~\cite{achlioptas2018learning,valsesia2018learning} in terms of fidelity and coverage, the non-uniformity issue remains unsolved, as illustrated in Fig.~\ref{fig:uniform_loss} (a).
For a clearer visualization, we colorize the points based on their relative relationships on TreeGCN. It shows that points with distant relationships might clutter in the 3D space. Without a proper regularization, points of different branches would tend to form a Gaussian-like distribution, such that more points are gathered around the geometric center of an object or the joints between different semantic parts, resulting in highly non-uniform shapes.

\begin{figure}[t]
\vspace{-2mm}
\begin{center}
 \includegraphics[width=0.99\linewidth]{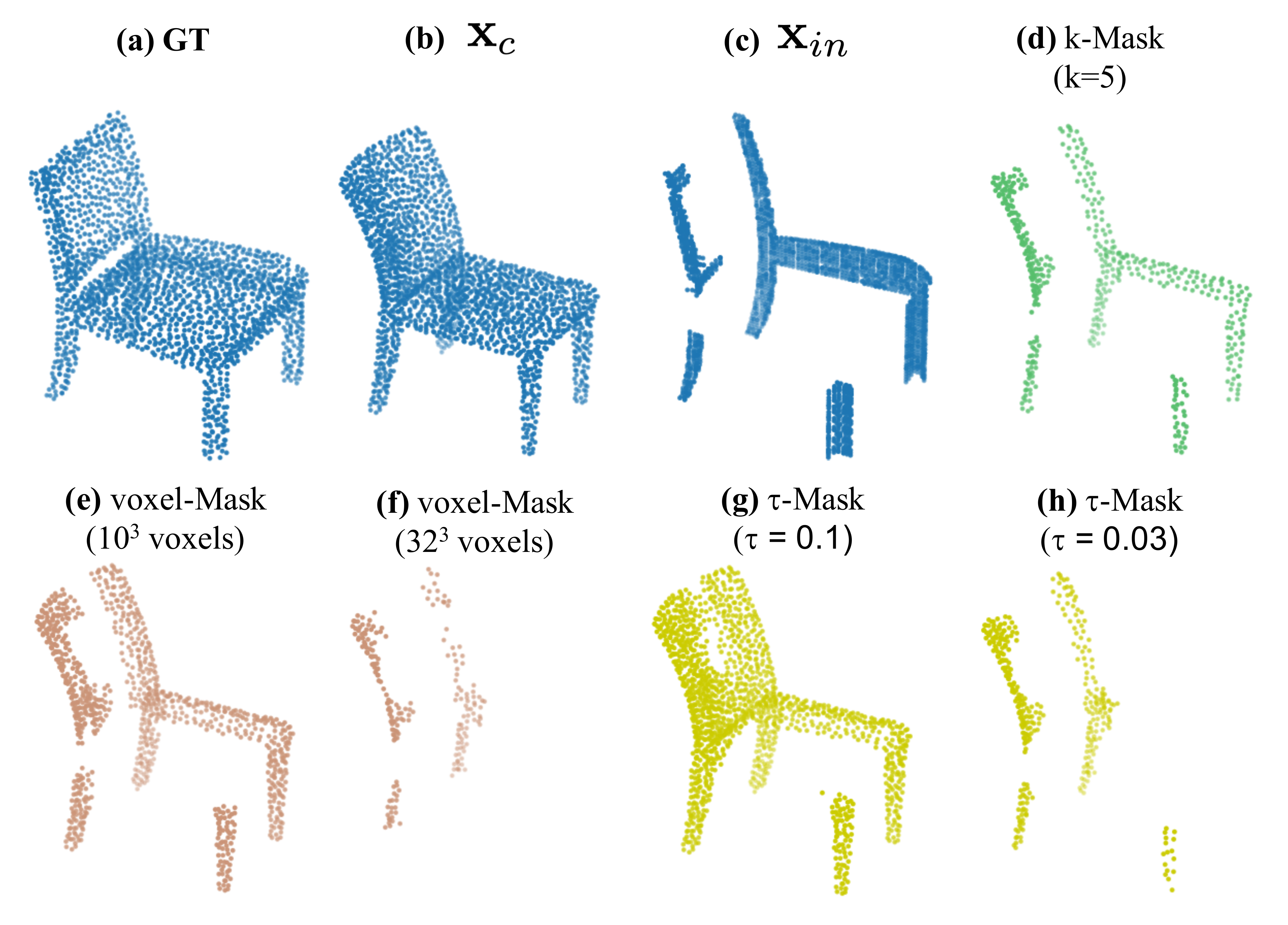}
\end{center}
\vspace{-5mm}
 \caption{Better degradation function produces partial shapes $x_{p}$ that are more similar to the input shape $x_{in}$. Our proposed k-Mask provides accurate point correspondence between $x_{in}$ and the generated complete shape $x_{c}$. In contrast, voxel-mask and $\tau$-mask are sensitive to the hyperparameters, \ie, the voxel size or the distance threshold respectively: large values result in noisy degradation, \eg, the chair's back has excessive points in \textbf{(e)} and \textbf{(g)}; small values lead to missing parts when correpsonding semantic parts reside differently in the 3D space, \eg, the chair's legs are falsely masked off in \textbf{(f)} and \textbf{(h)}. Note that $\vx_{in}$ has 2048 points in ShapeNet benchmark, while k-Mask is shown to be robust to partial shapes of any density}
 \vspace{-3.0mm}
\label{fig:masks}
\end{figure}

The non-uniformity of shapes' point clouds is a long-standing problem. Studies in point cloud upsampling~\cite{li2019pu,yu2018pu} propose some forms of uniformity losses, such as \textit{repulsion loss}, on point cloud patches. Besides, MSN~\cite{liu2019morphing} proposes \textit{expansion penalty} to reduce overlapping of the surface elements. However, these methods regularize each part of the shape separately, without enforcing a consensus across all parts to achieve an overall uniformity.
In view of their weaknesses, we propose a new uniform loss, \textbf{PatchVariance}, to regularize the uniformity of the entire shape during tree-GAN training, in addition to its adversarial loss.
\begin{equation}
\label{eq:patch}
  \cL_{patch} = Var(\{\rho _j\}_{j=1}^{n}), \quad \rho_{j} = \frac{1}{k} \Sigma_{i=1}^{k}dist_{ij}^{2}
\end{equation}

Specifically, we randomly sample $n$ seed positions over the object surface via Farthest Point Sampling (FPS), and then form small patches by including the $k$-nearest neighbors for each seed. Regardless of fine or bulk structure, these small patches shall scatter similarly. Thus we compute the average distance between each seed and its $k$-nearest neighbors and penalize the variance of all patches' average distances, as shown in Eq.~\eqref{eq:patch}.

As illustrated in Fig.~\ref{fig:uniform_loss} (b), PatchVariance significantly improves the uniformity of the generated shape. More evaluation and comparison with other uniform losses is covered in the ablation study.
Note that the proposed uniform loss is generic: it directly works on the generated shape and is invariant to the GAN architecture. Cross-validation of PatchVariance on r-GAN~\cite{achlioptas2018learning} (with MLP-based generator) can be found in the Supplementary Materials.

\subsection{Degradation in the 3D Space}
\label{sec:mask}

For shape completion, we define a degradation transform $M$ to best approximate the transformation from a generated complete shape $\vx_{c}=G(z)$ to a partial shape $\vx_{p}$, such that the corresponding regions between $\vx_{c}$ and the given partial shape $\vx_{in}$ can be precisely compared. We find that defining such a degradation function is ill-posed due to the unique unstructured nature of point clouds.

One may intuitively relate it to the image inpainting task, where a 2D binary mask $m$ is typically provided to degrade a complete image to the observation space through element-wise product: $x_{masked} = x \odot m$, given that the pixel correspondences between any image pair are consistent with the pixel locations. In contrast, corresponding regions of two 3D structures may reside at different locations in the 3D space, while directly voxelizing $\vx_{in}$ to form a 3D tensor indicating the voxel occupancy would inevitably lead to information loss. More importantly, as it is likely that $\vx_{c}$ is quite different from $\vx_{in}$ especially in the early GAN inversion stage, the corresponding semantic parts do not fall in the same voxels, hence leading to problematic degradation. See Fig.~\ref{fig:masks} (e) and (f) for an illustration.

In this work, we introduce an accurate and robust mask as the degradation function $M$, which we refer to as \textbf{k-Mask}.
An accurate and robust degradation shall be based on the knowledge of corresponding points between $\vx_{in}$ and a general $\vx_{c}$. In fact, point correspondences are ambiguous, far less straightforward, and variant to different generated shapes.
To this end, we dynamically obtain the point correspondences between $\vx_{in}$ and a specific $\vx_{c}$ based on the Euclidean distance. In view of the correspondence ambiguity, we opt for multiple corresponding points for a robust design. Specifically, for each point $p_i$ in $\vx_{in}$, we look for its k-nearest neighbors from $\vx_{c}$, denoted as $N_k^{\vx_{c}}(p_i)$. Consequently, $\vx_{p}$ can be constructed by the union of these k-nearest neighbors, as shown in Eq.~\eqref{eq:k-mask}. 

\begin{equation}
\label{eq:k-mask}
 \vx_{p} = \bigcup_{i=1}^n \{q_j \in N_k^{\vx_{c}}(p_i) \ | \  p_i \in \vx_{in} \}
\end{equation}

\noindent
\textbf{Alternative Design Variants}. We also provide other alternative masks for comparison. As stated above, the \textbf{voxel-Mask} is an intuitive design that directly extends the 2D binary mask to the 3D domain. Voxelization of $\vx_{in}$ gives its voxel occupancy, such that $\vx_{p}$ simply consists of all the points of $\vx_{c}$ that correspond to the occupied voxels in $\vx_{in}$. \textbf{$\tau$-Mask} determines the corresponding regions based on a predefined threshold. Eq.~\eqref{eq:tau-mask} describes $\vx_{p}$, which consists of all points of $\vx_{c}$ where the L2 distance from its nearest neighbor is within a threshold $\tau$.
\begin{equation}
\label{eq:tau-mask}
    \vx_{p} = \{q \in \vx_{c} \ | \ \min_{p \in \vx_{in}}{||p-q||_2} < \tau \}
\end{equation}

As illustrated in Fig.~\ref{fig:masks}, k-Mask provides an accurate and robust degradation whereas the other masks fail to achieve these two goals concurrently.
This is because both voxel-Mask and $\tau$-Mask leverage essentially fixed distance thresholds and are thus unable to adapt to changes in point density in certain regions. This observation is in line with the preference of k-NN over ball query in popular point feature extractors~\cite{wang2019dynamic, li2018pointcnn}.

\setlength{\tabcolsep}{8.3pt}
\begin{table*}
\vspace{-2mm}
\small
\caption{Effectiveness of PatchVariance on the shape uniformity. PatchVariance achieves the lowest MMD-EMD $\downarrow$ (scaled by $10^3$) across all the eight categories from ShapeNet, indicating the best uniformity and fidelity for the generated shapes}
\vspace{-2mm}
\label{tab:uniform}
\begin{center}
\begin{tabular}{l|c|c|c|c|c|c|c|c|c}
\hline
Methods   & Plane &  Cabinet & Car & Chair & Lamp  & Sofa & Table &   Boat   & Average  \\ 
\hline
tree-GAN baseline &	30.7 & 52.9 & 38.4 & 58.6 & 59.6 & 41.2 & 57.1 & 42.9 & 47.7 \\
tree-GAN + expansion penality~\cite{liu2019morphing} & 39.7 & 68.7 & 41.0 & 59.3 & 66.7 & 55.4 & 66.5 & 40.3 & 54.7 \\
tree-GAN + repulsion loss~\cite{yu2018pu} & 29.8 & 54.5 & 36.9 & 53.2 & 61.3 & 44.9 & 56.1 & 40.7 & 47.2 \\
tree-GAN + PatchVariance (ours) & \textbf{28.1} & \textbf{35.0} & \textbf{30.9} & \textbf{45.9} & \textbf{52.1} & \textbf{35.5} & \textbf{47.7} & \textbf{36.9} & \textbf{39.0} \\

\hline
\end{tabular}
\end{center}
\vspace{-8mm}
\end{table*}

\subsection{Loss Function for Inversion}
\label{sec:loss}
Chamfer Distance (CD) and Earth Mover's Distance (EMD) are the most commonly used structural losses for shape completion, with the latter being more sensitive to details and the density distribution~\cite{liu2019morphing}. However, unlike typical training processes of supervised shape completion that measure the distance between two complete shapes, our GAN inversion process compares a specific degraded shape against the given partial shape, which may contain a different number of points, thus making EMD infeasible. We follow the CD-T variant~\cite{wang2020cascaded,tchapmi2019topnet} that computes the squared L2 distance, as shown in Eq.~\eqref{eq:cd}.

\begin{equation}
\label{eq:cd}
\begin{aligned}
    \cL_{CD}(\vx_p, \vx_{in}) = \frac{1}{|\vx_{p}|} \sum_{p \in \vx_p} \min_{q \in \vx_{in}} ||p-q||_2^2 \\
    + \frac{1}{|\vx_{in}|} \sum_{q \in \vx_{in}} \min_{p \in \vx_p} ||p-q||_2^2
\end{aligned}    
\end{equation}

As structural losses are typically only concerned about low-level regularity of the point cloud,
we also perform feature matching at the observation space hoping to align the geometries more semantically. Following the recent practice in~\cite{pan2020exploiting}, we make use of the discriminator, a network that is trained together with the generator during pre-training. We take the feature from the intermediate layer immediately after max-pooling, which captures more geometric details, and compute the L1 distance as the \textbf{Feature Distance} loss, as shown in Eq.~\eqref{eq:d}.
\begin{equation}
\label{eq:d}
     \cL_{FD} =|| D(\vx_{p}) - D(\vx_{in})||_{1}
\end{equation}

The overall loss function is shown in Eq.~\eqref{eq:loss}, which is used in both shape completion and reconstruction of complete shapes.
\begin{equation}
\label{eq:loss}
    \cL = \cL_{CD} (\vx_{p}, \vx_{in}) + \cL_{FD}(\vx_{p}, \vx_{in})
\end{equation}


\section{Experiments}

We start with an ablation study (Sec.~\ref{subsec:ablation_study}) and then evaluate ShapeInversion through extensive experiments. Besides shape completion on the virtual scan benchmark (Sec.~\ref{subsec:virtual_scans}), we also compare its generalization with other methods on cross-domain partial shapes (Sec.~\ref{subsec:generalization_tests}) and real-world partial scans (Sec.~\ref{subsec:real_scans}). In addition, we also provide qualitative results on multiple valid output under ambiguity (Sec.~\ref{subsec:multiple_valid_outputs}) and shape manipulation of completed shapes (Sec.~\ref{subsec:shape_manipulation}).

\noindent
\textbf{Datasets}. To facilitate a comprehensive evaluation, we conduct experiments on both synthetic and real-world partial shapes. The following three forms of synthetic partial shapes are: \textbf{a)} virtual scans (\eg, in PCN~\cite{yuan2018pcn} and CRN~\cite{wang2020cascaded}) \textbf{b)} ball-holed partial shapes (\eg, in PF-Net~\cite{huang2020pf})  and \textbf{c)} semantic part-level incompleteness (PartNet~\cite{mo2019partnet}), as shown in Fig.~\ref{fig:teaser} (a)-(c). They are all derived from ShapeNet~\cite{chang2015shapenet}. For real-world scans, we evaluate on objects extracted from three sources: \textbf{i)} KITTI (cars)~\cite{geiger2012we}, \textbf{ii)} ScanNet (chairs and tables)~\cite{dai2017scannet}, and \textbf{iii)} MatterPort3D (chairs and tables)~\cite{chang2017matterport3d}, as shown in Fig.~\ref{fig:teaser} (d). Note that we follow the standard practice in the field of shape completion to assume the input is always canonically oriented.

\noindent
\textbf{Evaluation Metrics}. In Sec.~\ref{subsec:ablation_study}, we evaluate the fidelity and uniformity of the set of generated shapes against those in the test set using \textit{Minimum Matching Distance-Earth Mover's Distance (MMD-EMD)}~\cite{achlioptas2018learning, shu20193d}. 
EMD is highly indicative of uniformity as it conducts bijective matching of points between two point clouds.
With ground truth in Sec.~\ref{subsec:virtual_scans} and Sec.~\ref{subsec:generalization_tests}, we evaluate the shape completion performance using \textit{CD} and \textit{F1} score following pcl2pcl~\cite{chen2019unpaired}, where F1 is the harmonic average of the \textit{accuracy} and the \textit{completeness}. Without ground truth in Sec.~\ref{subsec:real_scans}, we use \textit{Unidirectional Chamfer Distance (UCD)} and \textit{Unidirectional Hausdorff Distance (UHD)}~\cite{chen2019unpaired,wu2020multimodal} from the partial input $\vx_{in}$ to the generated shape $\vx_{c}$.

\noindent
\textbf{Implementation Details}. In all experiments, ShapeInversion uses the same tree-GAN that is pre-trained on the ShapeNet train set for complete shape generation.
Although tree-GAN is able to generate multi-class 3D point clouds, we follow pcl2pcl and MPC~\cite{wu2020multimodal}, and train single-class models for each class for better fidelity.
The resolution of the predicted complete shape is 2048 for all the following experiments.
More details can be found in the Supplementary Materials.

\subsection{Ablation Study}
\label{subsec:ablation_study}
We first investigate the merit of each module in our framework, covering both the pre-training and the GAN inversion stage.

\begin{figure}
\begin{center}
\vspace{-2mm}
  \includegraphics[width=0.95\linewidth]{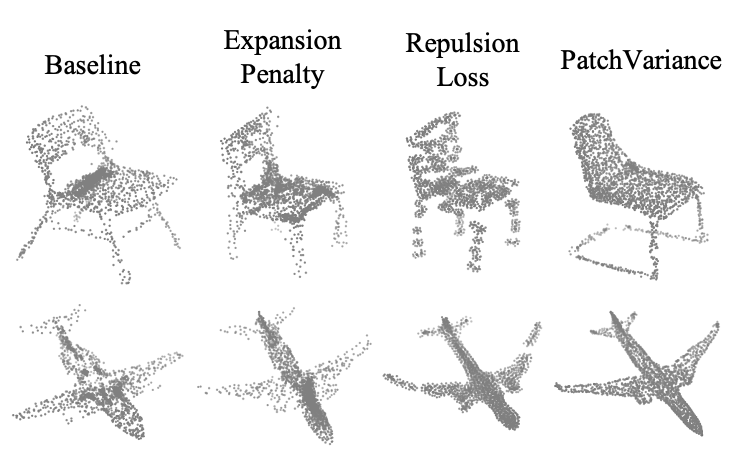}
\end{center}
\vspace{-4mm}
  \caption{Visualization of randomly generated shapes using various methods for uniformity. Each generated shape contains 2048 points, where darker regions indicates higher point density. PatchVariance achieves the best uniformity}
\label{fig:uniform_results}
\vspace{1mm}
\end{figure}

\noindent
\textbf{Effectiveness of PatchVariance}. We compare our PatchVariance against expansion penalty~\cite{liu2019morphing} and repulsion loss~\cite{yu2018pu}. As shown in Tab.~\ref{tab:uniform}, PatchVariance achieves the best result across all categories. From Fig.~\ref{fig:uniform_results}, we can observe that expansion penalty leads to more unevenly distributed point clouds while it penalizes branch expansion, and repulsion loss enforces uniformity at local regions only whereas a global uniformity is obtained with PatchVariance.

\noindent
\textbf{Effectiveness of k-Mask and Feature Distance}. Tab.~\ref{tab:albation_inversion} shows the ablation study during the GAN inversion stage. Replacing k-Masks with other alternative degradation functions shows significant degradation. The choice of k-nearest neighbors of points in the partial shape provides an accurate and robust degradation, and better adapts to variations in local point density. The use of feature distance provides more semantic information to complement the structural loss, significantly boosting the performance.
 
\setlength{\tabcolsep}{4.5pt}
\begin{table}
\small
\caption{Effectiveness of various degradation functions and Feature Distance. Note that the results with these masks are obtained at their respective optimal hyperparameters: $10^3$ voxels for voxel-Mask, $\tau=0.03$ for $\tau$-Mask, and k $=$ 5 for k-Mask}
\label{tab:albation_inversion}
\vspace{-2mm}
\begin{center}
\begin{tabular}{l|c|c|c|c}
\hline
Methods   & CD ($\times10^4$) $\downarrow$ &  acc. $\uparrow$ & comp. $\uparrow$ & F1 $\uparrow$  \\ 
\hline
Ours w/ voxel-Mask & 19.3 &	84.7 &	79.7 &	81.5 \\
Ours w/ $\tau$-Mask & 18.9 &	82.9 &	81.6 &	81.6 \\
Ours w/o $\cL_{FD}$ & 16.3 & 83.6 &	81.7 &	81.9 \\
Ours & \textbf{14.9} &	\textbf{85.0}	& \textbf{84.0}	& \textbf{83.9} \\
\hline
\end{tabular}
\end{center}
\vspace{-9mm}
\end{table}

\setlength{\tabcolsep}{4.5pt}
\begin{table*}
\vspace{-1mm}
\small
\caption{Shape completion results on ShapeNet benchmark. The numbers shown are [CD $\downarrow$ / F1 $\uparrow$], where CD is scaled by $10^4$. ShapeInversion outperforms pcl2pcl by a large margin, and is comparable to the various supervised methods. \textit{sup.}: supervised methods; \textit{unsup.}: unsupervised methods }
\label{tab:benchmark}
\vspace{-2mm}
\begin{center}
\begin{tabular}{l|l|c|c|c|c|c|c|c|c|c}
\hline
 {} & Methods   & Plane &  Cabinet & Car & Chair & Lamp  & Sofa & Table &   Boat   & Average  \\   
\hline
\textit{sup.} & PCN~\cite{yuan2018pcn} &  3.5$/$96.5 & 11.3$/$86.4  & 6.4$/$94.0  & 11.0$/$86.0  & 11.6$/$84.6  &  11.5$/$85.2 &  10.4$/$89.4 & 7.4$/$91.7  &  9.1$/$89.2 \\
{} & TopNet~\cite{tchapmi2019topnet}  &  4.1$/$96.0  & 12.9$/$84.1  & 7.8$/$91.3  & 13.4$/$82.3  &  14.8$/$79.4 &  16.0$/$80.8 & 12.9$/$85.7  &  8.9$/$89.3 &  11.4$/$86.1 \\
{} & MSN~\cite{liu2019morphing}  &  2.9$/$97.4  & 12.5$/$85.5  & 7.1$/$92.3  &  10.6$/$86.8 &  9.3$/$88.6 & 12.0$/$83.3  &  9.6$/$91.3 &  6.5$/$93.1 &  8.8$/$89.8 \\
{} & CRN~\cite{wang2020cascaded}  & 2.3$/$98.3   &  11.4$/$86.2 & 6.2$/$93.8  & 8.8$/$89.7  &  8.5$/$90.2 & 11.3$/$85.1  & 9.3$/$92.9  & 6.1$/$94.2  &  8.0$/$91.3 \\
\hline
\textit{unsup.} & pcl2pcl~\cite{chen2019unpaired}  & 9.8$/$89.1   &  27.1$/$68.4  & 15.8$/$80.0  & 26.9$/$70.4  & 25.7$/$70.4  & 34.1$/$58.4  & 23.6$/$79.0  & 15.7$/$77.8  &  22.4$/$74.2 \\
{} & Ours  &  \textbf{5.6}$/$\textbf{94.3}  & \textbf{16.1}$/$\textbf{77.2}  &  \textbf{13.0}$/$\textbf{85.8} & \textbf{15.4}$/$\textbf{81.2}  &  \textbf{18.0}$/$\textbf{81.7}  & \textbf{24.6}$/$\textbf{78.4}   & \textbf{16.2}$/$\textbf{85.5}  &  \textbf{10.1}$/$\textbf{87.0} & \textbf{14.9}$/$\textbf{83.9 } \\
\hline
\end{tabular}
\end{center}
\vspace{-5mm}
\end{table*}
\setlength{\tabcolsep}{1.4pt}

\subsection{Shape Completion on Virtual Scan Benchmark}
\label{subsec:virtual_scans}
We compare with existing supervised and unsupervised methods on the common virtual scan benchmark generated from ShapeNet, first proposed by PCN~\cite{yuan2018pcn}. For a fair comparison, all the baseline methods are trained with virtual scans provided by CRN~\cite{wang2020cascaded} (with corresponding complete shapes from ShapeNet train set).
Tab.~\ref{tab:benchmark} shows that ShapeInversion outperforms the other unsupervised method pcl2pcl by a large margin across all the eight categories, and is comparable to the various supervised methods.
Note that the impressive performance of various supervised methods is in part attributed to the coarse-to-fine strategy, some of which even calibrates the coarse output with the partial input during the refining stage \cite{liu2019morphing,wang2020cascaded}. ShapeInversion, in contrast, performs completion in a single stage and achieves comparable results.
Besides Fig.~\ref{fig:teaser} (a), more qualitative results can be found in the Supplementary Materials.

\subsection{Robustness to Varying Partial Forms}
\label{subsec:generalization_tests}

\begin{figure}[t]
\begin{center}
  \includegraphics[width=0.99\linewidth]{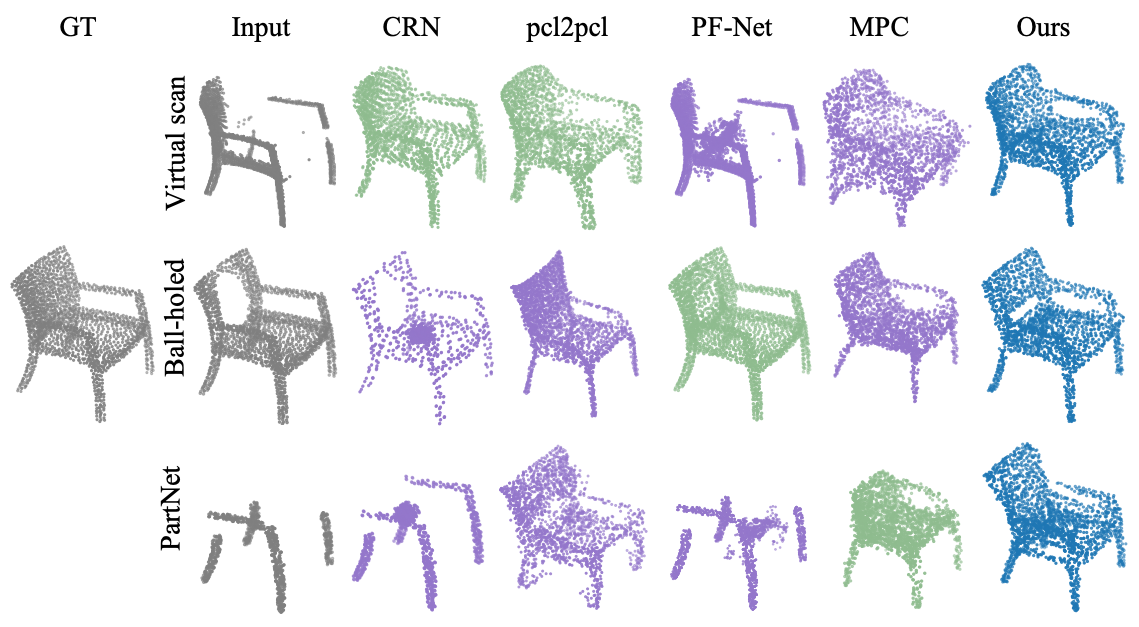}
  \vspace{-5mm}
\end{center}
  \caption{Visualization of cross-domain validation. Different partial forms of the same object are tested. In-domain results are in green whereas out-of-domain ones are in purple. Supervised methods, CRN and PF-Net, show significant performance drop with domain change; unsupervised methods pcl2pcl and MPC show relatively better results for out-of-domain inputs. In contrast, ShapeInversion constantly provides plausible and accurate outputs for all partial forms. Note that CRN leverages the partial input during the refinement stage; PF-Net only predicts the missing regions and combines the partial input as the final output}
\label{fig:cross_data}
\vspace{-2.5mm}
\end{figure}

To mimic various causes of partial shapes such as occlusion and self-occlusion, various partial forms such as ball-holed partial shapes and virtual scans are considered in different works. We demonstrate the robustness of ShapeInversion under three different partial forms in Tab.~\ref{tab:cross_data}. 
Supervised methods may bias towards the partial forms seen in the training pairs and in turn give poor results on out-domain data, even with auxiliary adversarial loss (\eg, CRN). The unsupervised pcl2pcl performs better than the supervised methods.
For ShapeInversion, the GAN is pre-trained with complete shapes only, and the degradation via k-Mask during the inversion stage is invariant to partial form changes. In this way, ShapeInversion achieves the best results across almost all the domains.
See Fig.~\ref{fig:cross_data} and Fig.~\ref{fig:teaser} (a)-(c) for qualitative results.

Note that PF-Net is trained to generate missing regions only for the ball-holed partial shape, which is not compatible with other partial forms with multiple missing regions; although MPC is able to give multiple outputs in view of ambiguity in the partial shape, we report results from its single output for a fair comparison. 
To further ensure fairness, we remove the shapes from the PartNet test split that are present in the ShapeNet train set. 

\begin{figure*}[t]
\vspace{-2mm}
\begin{center}
  \includegraphics[width=0.99\linewidth]{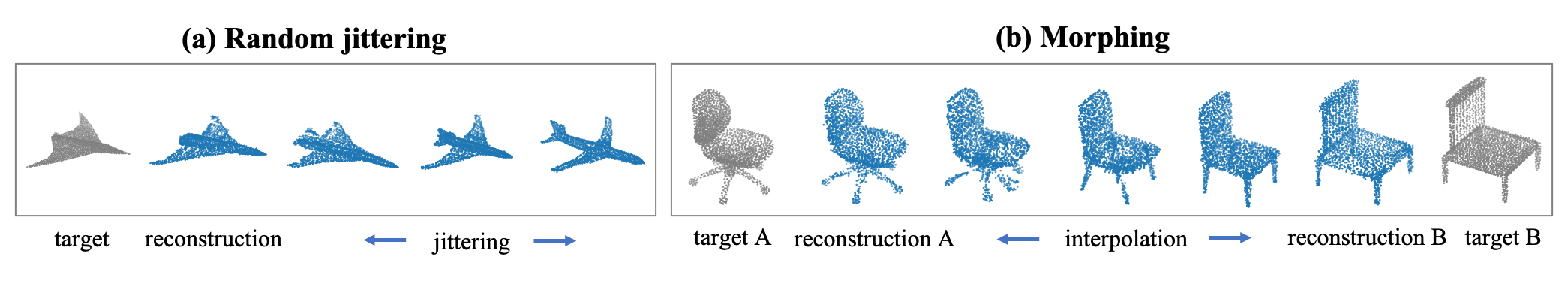}
\end{center}
\vspace{-6mm}
  \caption{ShapeInversion enables manipulation of complete shapes: \textbf{(a)} changing an object into other plausible shapes of different geometries; \textbf{(b)} making a sound transition from one shape to another}
  \vspace{-3mm}
\label{fig:manipulation}
\end{figure*}
\setlength{\tabcolsep}{3.5pt}
\begin{table}
\small
\caption{Cross-domain validation. We follow the literature to train each method on a certain partial form (\textit{source}) and cross-validate on other partial forms (\textit{targets}). For each target domain, the SOTA in-domain results are listed at the first line for reference. Methods, especially supervised ones, usually perform well on the in-domain data but suffer large performance drops on the out-of-domain data, whereas ShapeInversion gives the best results for almost all the cross-domain tests, highlighting its robustness to partial form changes. The metric is CD$\downarrow$ ($\times 10^4$)} 
\label{tab:cross_data}
\vspace{-2mm}
\begin{center}
\begin{tabular}{l|c|c|c|c|c}
\hline
Target & Methods & Source & Chair & Table & Lamp \\
\hline
\multirow{3}{*}{Virtual scan} & CRN	 & Virtual scan  &	8.8  &	9.3  &	8.5 \\ \cline{2-6}
{} & MPC~\cite{wu2020multimodal}  &	PartNet  &	45.9  &	88.9  &	63.0 \\
{} & Ours  &	-	 & \textbf{15.4}  &	\textbf{16.2} &	\textbf{18.0} \\
\hline
\multirow{6}{*}{Ball-holed}	  & PF-Net~\cite{huang2020pf}  &	Ball-holed  &	11.9  &	9.9  &	23.1 \\ \cline{2-6}
{} & 	MSN  &	Virtual scan  &	79.6  &	46.6  &	55.4 \\
{} & 	CRN	  &Virtual scan	  & 44.7  &	52.9  &	52.1 \\
{} & 	pcl2pcl  &	Virtual scan  &	18.6  &	18.5  &	21.2 \\
{} & 	MPC  &	PartNet	  & 44.7	  & 28.9  &	69.5 \\
{} & Ours  &	-	 & \textbf{10.1}	 & \textbf{16.0} & \textbf{17.3} \\
\hline
\multirow{5}{*}{PartNet} & MPC &	PartNet &	40.0 &	51.0 &	82.0 \\ \cline{2-6}
 {} & MSN &	Virtual scan &	198.0 &	143.2 &	229.9 \\
{} & CRN &	Virtual scan &	177.4 &	140.6 &	185.9 \\
 {} & pcl2pcl &	Virtual scan &	51.0 &	\textbf{76.6} &	111.2 \\
 {} & Ours  &	- &	\textbf{36.8} &	77.8 &	\textbf{100.8} \\
\hline
\end{tabular}
\end{center}
\vspace{-5mm}
\end{table}

\subsection{Completion of Real-World Scans} 
\label{subsec:real_scans}

\begin{figure}[t]

\begin{center}
  \includegraphics[width=0.99\linewidth]{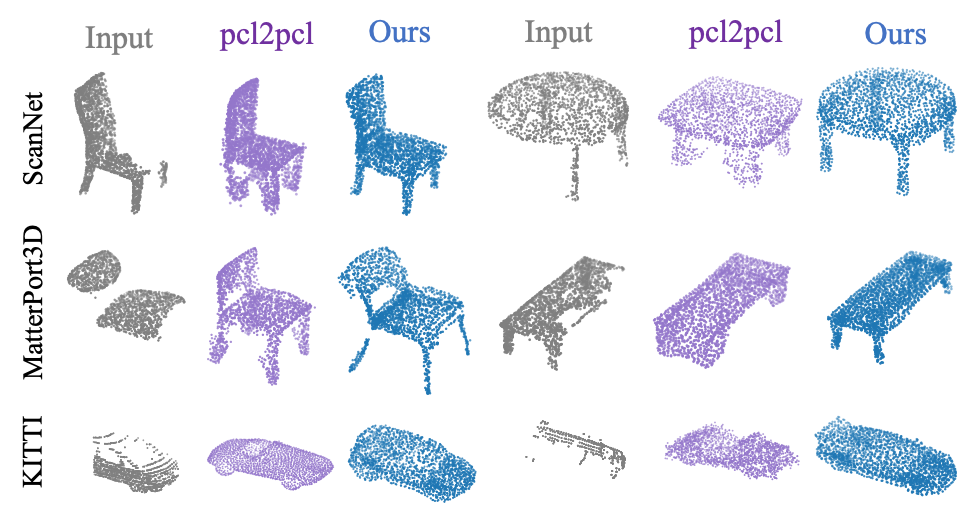}
\end{center}
\vspace{-5mm}
  \caption{Shape completion on real-world partial scans. Note that pcl2pcl is retrained with real-world partial shapes (together with synthetic complete shapes in an unpaired manner~\cite{chen2019unpaired}). In contrast, ShapeInversion does not use any real scans, yet, reconstructs high-fidelity shapes that are more faithful to the partial input}
\label{fig:realscan}
\vspace{-3mm}
\end{figure}

We investigate the generalization of ShapeInversion further on real-world data extracted from MatterPort3D, ScanNet, and KITTI. Besides the domain gap from virtual scans, these real scans tend to be noisier and more incomplete, \eg, KITTI cars. We quantitatively evaluate the performance of ShapeInversion and pcl2pcl in Tab.~\ref{tab:realscan} using UCD and UHD. Despite that pcl2pcl is retrained with real-world scans, our approach significantly outperforms pcl2pcl in terms of UCD and achieves comparable UHD given that pcl2pcl is trained via UHD. With the further addition of UHD into the loss function, ShapeInversion achieves better UHD results with a small compromise on the UCD performance. 
The completion results in Fig.~\ref{fig:realscan} reveals that pcl2pcl tends to ignore the geometric details in the partial shape, whereas our results remain highly plausible and faithful.

\setlength{\tabcolsep}{2.2pt}
\begin{table}
\caption{Shape completion results on the real scans. As there is no corresponding ground truth, we evaluate the results using [UCD $\downarrow$ / UHD $\downarrow$], where UCD is scaled by $10^4$ and UHD is scaled by $10^2$
}
\vspace{2mm}
\centering
\small
\begin{tabular}{l|c|c|c|c|c}
\hline

\multirow{2}{*}{Methods} & \multicolumn{2}{c}{ScanNet} & \multicolumn{2}{c}{MatterPort3D} & KITTI \\
\cline{2-6}
{}& Chair & Table & Chair & Table & Car \\
\hline
pcl2pcl & 17.3$/$10.1 & 9.1$/$11.8 & 15.9$/$10.5 & 6.0$/$11.8 & 9.2$/$14.1 \\
\hline
Ours & \textbf{3.2}$/$10.1 & \textbf{3.3}$/$11.9 & \textbf{3.6}$/$10.0 & \textbf{3.1}$/$11.8 & \textbf{2.9}$/$13.8 \\
\hline
Ours+UHD & 4.0$/$\textbf{9.3} & 6.6$/$\textbf{11.0} & 4.5$/$\textbf{9.5} &  5.7$/$\textbf{10.7} & 5.3$/$\textbf{12.5} \\
\hline
\end{tabular}
\vspace{1mm}

\label{tab:realscan}
\end{table}

\subsection{Multiple Valid Outputs under Ambiguity}
\label{subsec:multiple_valid_outputs}

\begin{figure}[t]
\vspace{-1mm}
\begin{center}
  \includegraphics[width=0.99\linewidth]{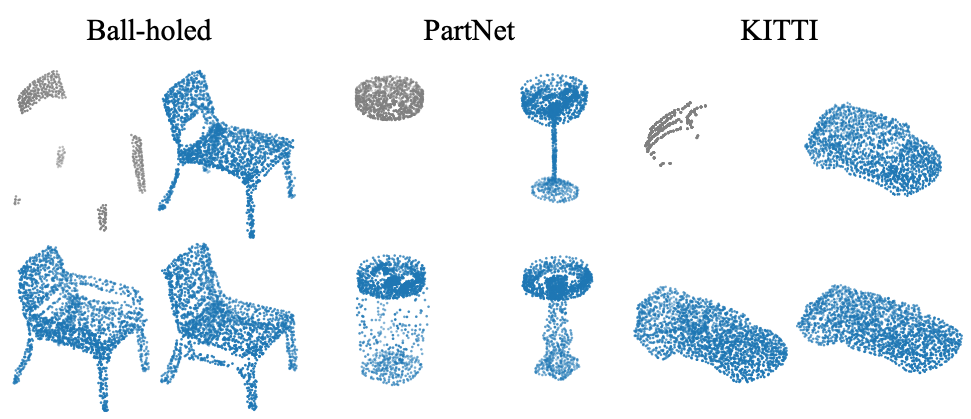}
  \vspace{-4mm}
\end{center}
  \caption{ShapeInversion can give multiple valid outputs when higher incompleteness level of partial shapes impose ambiguity}
\label{fig:multi_outputs}
\vspace{-4mm}
\end{figure}

With more severely incomplete input, there is more than one complete shape that makes sense. Our framework can naturally give multiple valid and diversified outputs, as we can inverse from multiple initial values of $z$, which are selected from hundreds of initial values via FPS, subject to the loss $\cL$ being smaller than a threshold $\tau_{\cL}$.
As demonstrated in Fig.~\ref{fig:teaser} (e) and Fig.~\ref{fig:multi_outputs}, ShapeInversion provides multiple reasonable outputs, where each of them faithfully reflects the details in the partial shape.

There exists a trade-off between the diversity and fidelity of the output shapes. In contrast to MPC~\cite{wu2020multimodal} where the trade-off is predefined during the training by the weights of different losses, our framework offers a more flexible diversity-fidelity trade-off, \eg, we can opt for higher diversity for a particular partial shape by simply choosing a large $\tau_{\cL}$ and reducing the number of iterations of inversion.

\subsection{Shape Manipulation}
\label{subsec:shape_manipulation}

Shape manipulation enables interesting applications such as generative design. We show that ShapeInversion can be readily extended to \textbf{random jittering} and \textbf{morphing}, giving plausible new shapes and sound transition from one shape to another respectively, as shown in Fig.~\ref{fig:manipulation}. These can be realized efficiently upon shape reconstruction: jittering of a given shape is achieved by introducing perturbation in the latent space; morphing between two given shapes is achieved by interpolation between their corresponding latent vectors $\vz$ and generator parameters $\theta$.

\section{Conclusion}

We introduce ShapeInversion for unsupervised point cloud completion. ShapeInversion addresses the domain gaps between virtual and real-world partial scans, and among various simulated partial shapes through GAN inversion. 
As the very first GAN inversion approach for 3D shape completion, we introduce two new components to address the unique challenges posed by the nature of points clouds: an effective uniform loss, PatchVariance, and an accurate and robust degradation function, k-Mask.  
With the incorporation of rich knowledge of shape geometries and semantics captured in a well-trained GAN, it achieves remarkable generalization for real-world scans and partial inputs of various forms and incompleteness levels. Moreover, our framework also brings several side-benefits, including giving multiple reasonable complete shapes for one partial input, as well as shape jittering and shape interpolation.

So far, both shape completion and manipulation are conducted on a model pre-trained with a single category. Future works can focus on improving the fidelity of multi-class models, which could provide more possibilities such as cross-category shape completion via a conditional GAN.

\noindent
\textbf{Acknowledgements}. This research was conducted in collaboration with SenseTime. This work is supported by A*STAR through the Industry Alignment Fund - Industry Collaboration Projects Grant.

\appendix
\renewcommand\thesection{\Alph{section}}
\section{Additional Details of Evaluation Metrics}

\noindent
\textbf{Minimum Matching Distance (MMD)}~\cite{achlioptas2018learning} measures the fidelity of a set of generated shapes $\Sbb_{c}$ with respect to the set of ground truth shapes $\Sbb_{gt}$. Specifically, we match every shape $\vx_{gt}$ in $\Sbb_{gt}$ to the generated shape $\vx_c$ in $\Sbb_{c}$ with minimum distance, and report the average macthed distance.

\noindent
\textbf{Earth Mover's Distance (EMD)}~\cite{rubner2000earth} is the solution, \ie, the minimum cost, to transport one shape's point cloud to another. It is defined in Eq.~\eqref{eq:emd}, where $\phi$ is a bijection. The bijection is highly indicative of uniformity of the generated shapes. We use the implementation provided by MSN~\cite{liu2019morphing}. 
\begin{equation}
\label{eq:emd}
\begin{aligned}
    d_{EMD}(\vx_A, \vx_B) = \min_{\phi : \vx_A \rightarrow \vx_B}
    \frac{1}{|\vx_{A}|} 
    \sum_{p \in \vx_A} ||p-\phi(p)||_2
\end{aligned}    
\end{equation}

\noindent
\textbf{Unidirectional Chamfer Distance (UCD)} measures the squared L2 distance from the partial input shape $\vx_p$ to the complete output $\vx_c$, as shown in Eq.~\eqref{eq:ucd}.

\begin{equation}
\label{eq:ucd}
    d_{UCD}(\vx_p,\vx_c)  = \frac{1}{|\vx_{p}|} \sum_{p \in \vx_p} \min_{q \in \vx_c} ||p-q||_2^2 
\end{equation}

\noindent
\textbf{Unidirectional Hausdorff Distance (UHD)}~\cite{chen2019unpaired,wu2020multimodal}, similarly,  measures the single-sided Hausdorff distance:
\begin{equation}
    d_{UHD}(\vx_p,\vx_c)  =  \max_{p \in \vx_p} \min_{q \in \vx_c} ||p-q||_2^2 
\end{equation}

\noindent
\textbf{F1 score}, following pcl2pcl~\cite{chen2019unpaired}, is defined as the harmonic average of the \textbf{accuracy} and the \textbf{completeness}, where accuracy measures the fraction of points in the $\vx_c$ that are matched in the corresponding ground truth shape $\vx_{gt}$, and completeness measures the fraction of points in the $\vx_{gt}$ that are matched in $\vx_c$. Here, the status of matching is determined by a threshold $\epsilon=0.03$ for L2 distance. 

\section{Additional Details of Datasets}
%
In Sec. \textcolor{red}{4.3} of the main paper, both the virtual scans and the ball-holed partial shapes follow the same train-test split of ShapeNet~\cite{chang2015shapenet}, whereas the PartNet~\cite{mo2019partnet} follows the train-test split provided by MPC~\cite{wu2020multimodal}. As some shapes in the PartNet test set are present in the ShapeNet train set, we remove them for evaluating the PartNet target domain.
The ball-holed partial shapes in this section are generated as described in PF-Net~\cite{huang2020pf}, \ie, removing exactly 512 points from the complete shape.
But in Fig.~\textcolor{red}{1} and Fig.~\textcolor{red}{9} of the main paper, higher incompleteness levels are tested to assess the robustness of ShapeInversion.

The real-world partial scans are provided by pcl2pcl~\cite{chen2019unpaired}.
Both ScanNet~\cite{dai2017scannet} and KITTI~\cite{geiger2012we} are split into train set and test set, and the mapping GAN of the pcl2pcl framework, which maps the latent space of partial shapes to that of the complete ones, is retrained on the real scan train set. Note that MatterPort3D~\cite{chang2017matterport3d} does not have its own train set, so pcl2pcl is evaluated on MatterPort3D using the model retrained on ScanNet, given that both these two datasets are captured by depth cameras. In contrast, ShapeInversion does not need the real scan train set, and is directly evaluated on the test set.

\section{Additional Details on Implementation}
\noindent
\textbf{Pre-training}. We follow Shu \etal~\cite{shu20193d} to train the tree-GAN baseline. The PatchVariance module has $n=100$ patches, each with $k=30$ points. This would ensure the entire generated shape (with 2048 points) is sampled via FPS.
This setting works for r-GAN~\cite{achlioptas2018learning} as well, as shown in Fig.~\ref{fig:uniform_rgan}. 
tree-GAN is trained on eight Nvidia V100 GPUs for 2000 epochs. For the four categories with more than 3000 shapes, \ie, plane, car, chair, and table, the batch size is 512; for the other four categories with fewer shapes, \ie, cabinet, lamp, sofa, and boat, the batch size is 128 to train for enough iterations.

\noindent
\textbf{Inversion}. The $k$ in k-Mask is 5. We manually split the inversion stage into four sub-stages, each with different learning rates for $z$ and $\theta$: $\alpha_z = [ 1\times 10^{-2}, 1\times 10^{-4},1\times 10^{-5},1\times 10^{-6}] $, and $\alpha_{\theta} = [2\times 10^{-7},1\times 10^{-6},1\times 10^{-6},2\times 10^{-7}] $. For bulk structures, \ie, car, couch, cabinet, and plane,  each sub-stage consists of 30 iterations; for thin structures, \ie, chair, lamp, table, and boat, each sub-stage consists of 200 iterations.

\section{Additional Qualitative Results}

\begin{figure*}[t]
\begin{center}
  \includegraphics[width=0.99\linewidth]{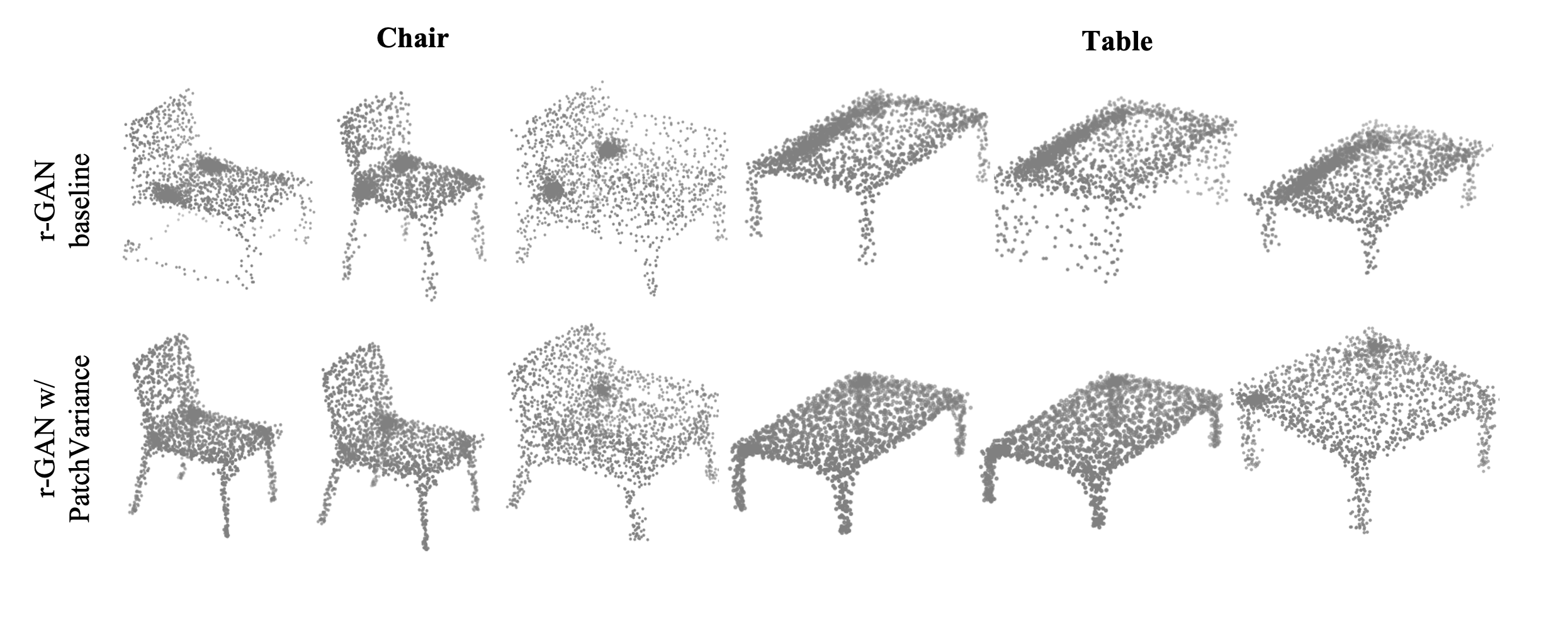}
  \vspace{1mm}
\end{center}
  \caption{ Validation of the significance of PatchVariance on r-GAN~\cite{achlioptas2018learning}. Non-uniformity issue is a common problem faced by point cloud GANs, our PatchVariance loss directly works on the generated point cloud, and is invariant to the architecture of the generator, hence it effectively enhances the uniformity of generated point clouds for r-GAN  as well. Darker regions have a higher density of points}
  \vspace{0mm}
\label{fig:uniform_rgan}
\vspace{3mm}
\end{figure*}

\begin{figure*}[t]
\vspace{3mm}
\begin{center}
  \includegraphics[width=0.99\linewidth]{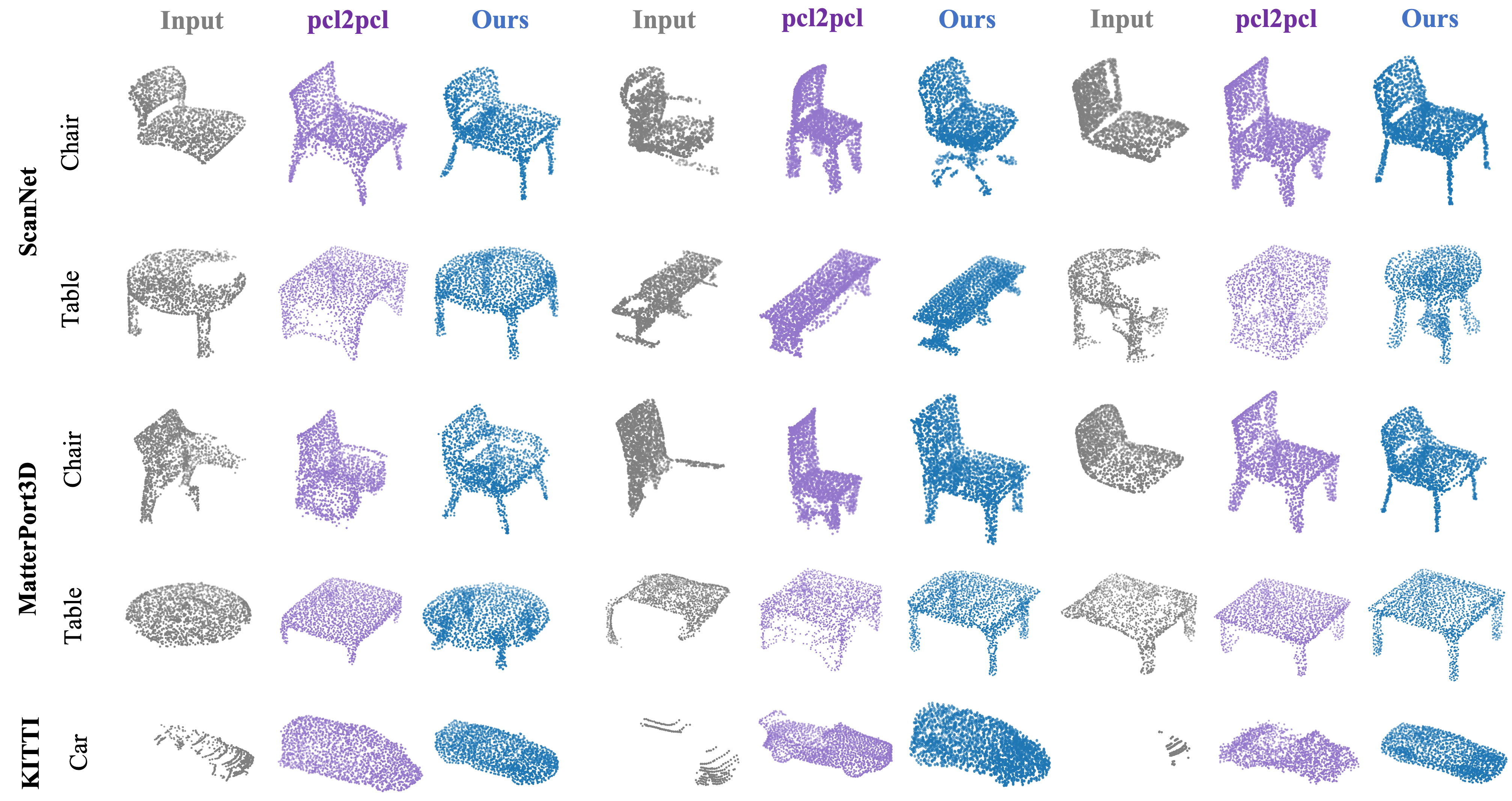}
\end{center}
\vspace{4mm}
  \caption{More qualitative results for shape completion on real-world partial scans. Note that pcl2pcl is retrained with real-world partial shapes (together with synthetic complete shapes in a unpaired manner~\cite{chen2019unpaired}).
  In contrast, ShapeInversion does not use any real scans, yet, reconstructs high-fidelity shapes that are more faithful to the partial input}
\label{fig:realscan2}
\vspace{8mm}
\end{figure*}

\begin{figure*}[t!]
\begin{center}
\vspace{8mm}
  \includegraphics[width=0.99\linewidth]{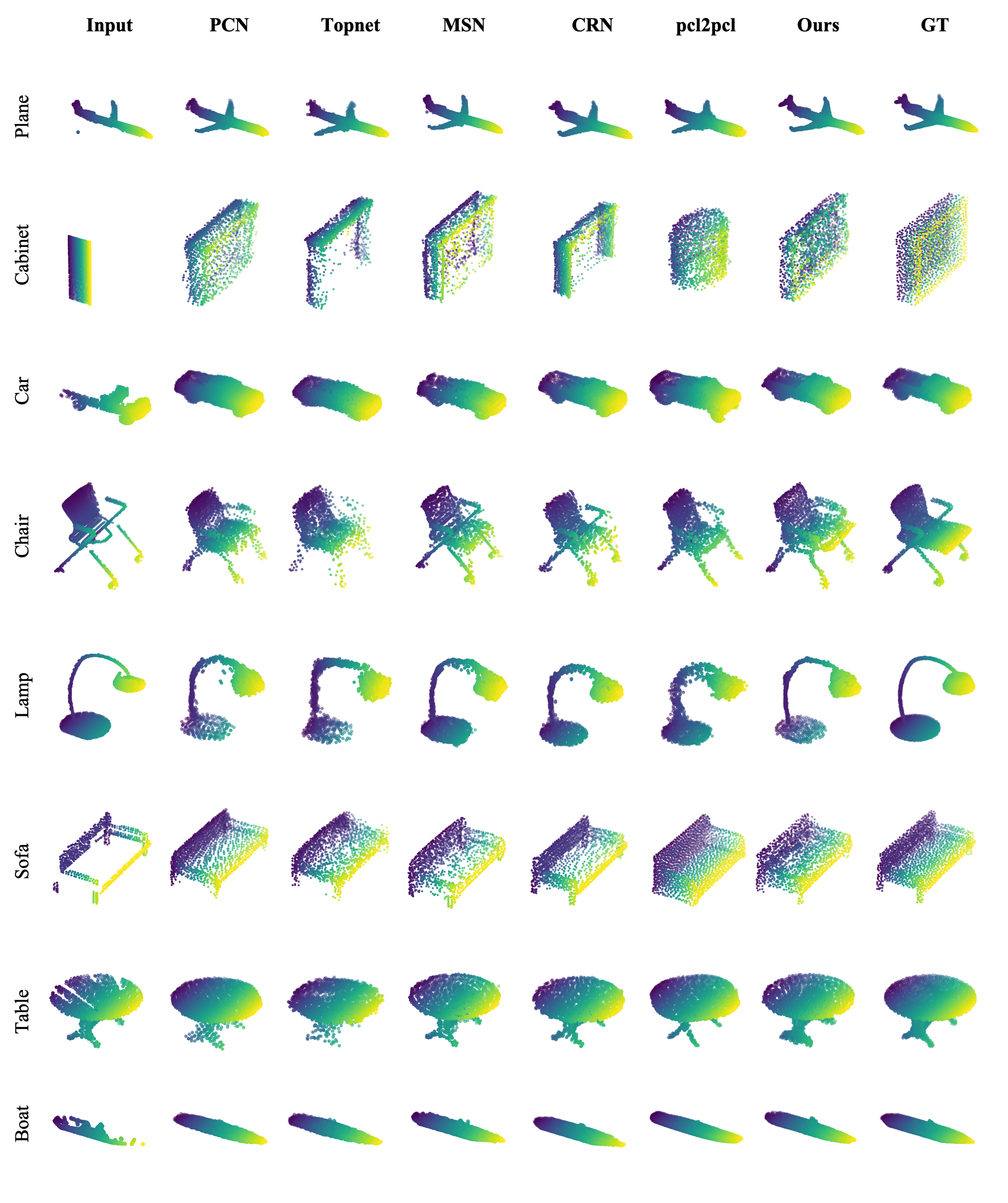}
  \vspace{-3mm}
\end{center}
  \caption{Qualitative comparison on the ShapeNet benchmark. The resolution of partial inputs, complete shapes, and ground truths are all 2048 points}
\label{fig:benchmark}
\vspace{8mm}
\end{figure*}

\begin{figure*}[t]
\begin{center}
\vspace{-1mm}
  \includegraphics[width=0.99\linewidth]{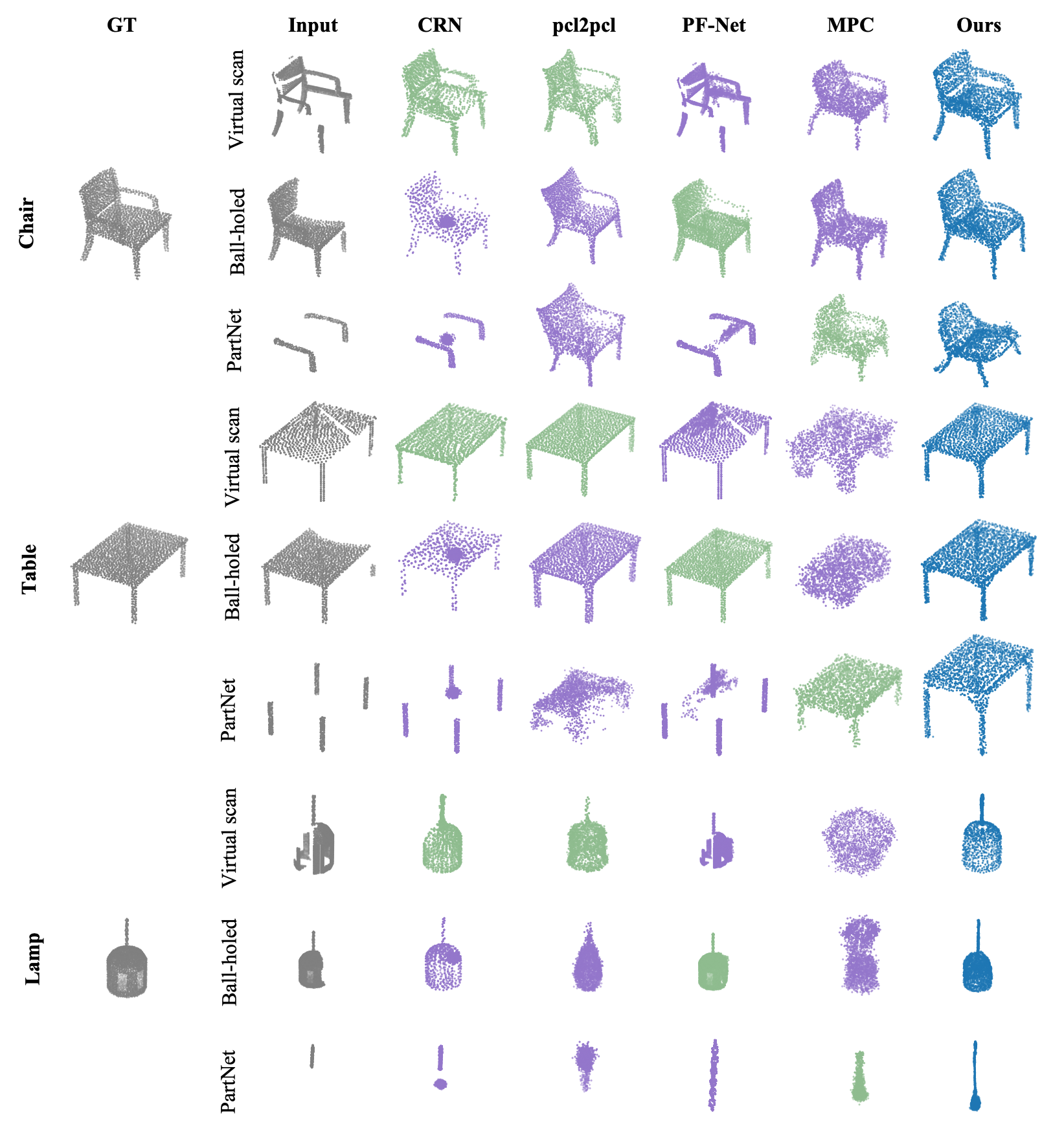}
  \vspace{-5mm}
\end{center}
  \caption{Additional visualization of cross-domain validation. Different partial forms of the same object are tested. In-domain results are in dark green whereas out-of-domain ones are in purple. Supervised methods CRN and PF-Net show significant performance drop with domain change; unsupervised methods pcl2pcl and MPC show relatively better results for out-of-domain inputs. ShapeInversion constantly provides plausible and accurate outputs for all partial forms. Note that CRN leverages the partial input during the refinement stage; PF-Net only predicts the missing regions and combines the partial input as the final output}
\label{fig:cross_data2}
\vspace{-2mm}
\end{figure*}

\begin{figure*}[t]

\begin{center}
  \includegraphics[width=0.95\linewidth]{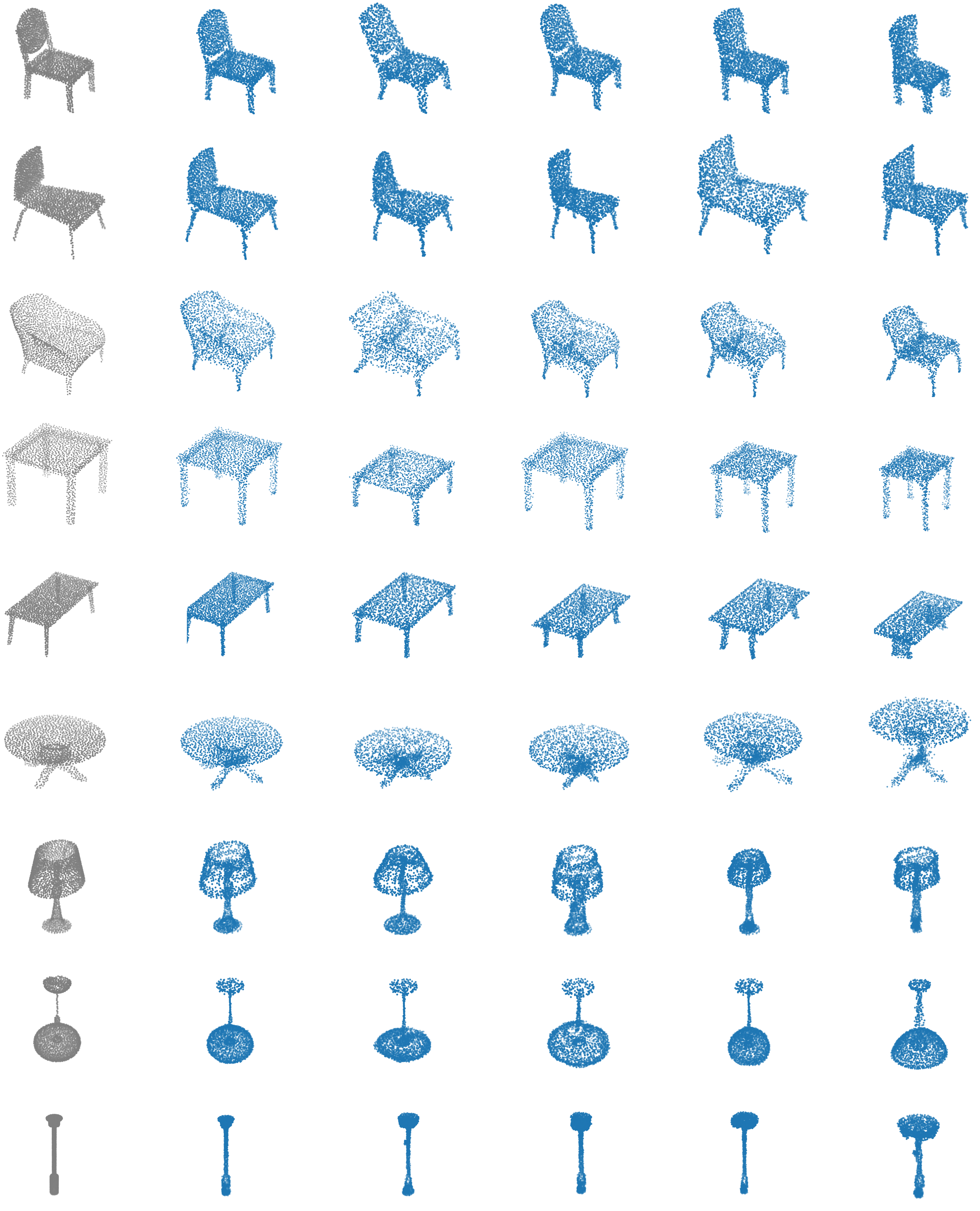}
\end{center}
  \caption{Additional qualitative results for shape random jittering by ShapeInversion. Random jittering changes one shape into other plausible shapes of different geometries by introducing perturbation in the latent space}
\label{fig:jit}
\end{figure*}

\begin{figure*}[t]

\begin{center}
  \includegraphics[width=0.99\linewidth]{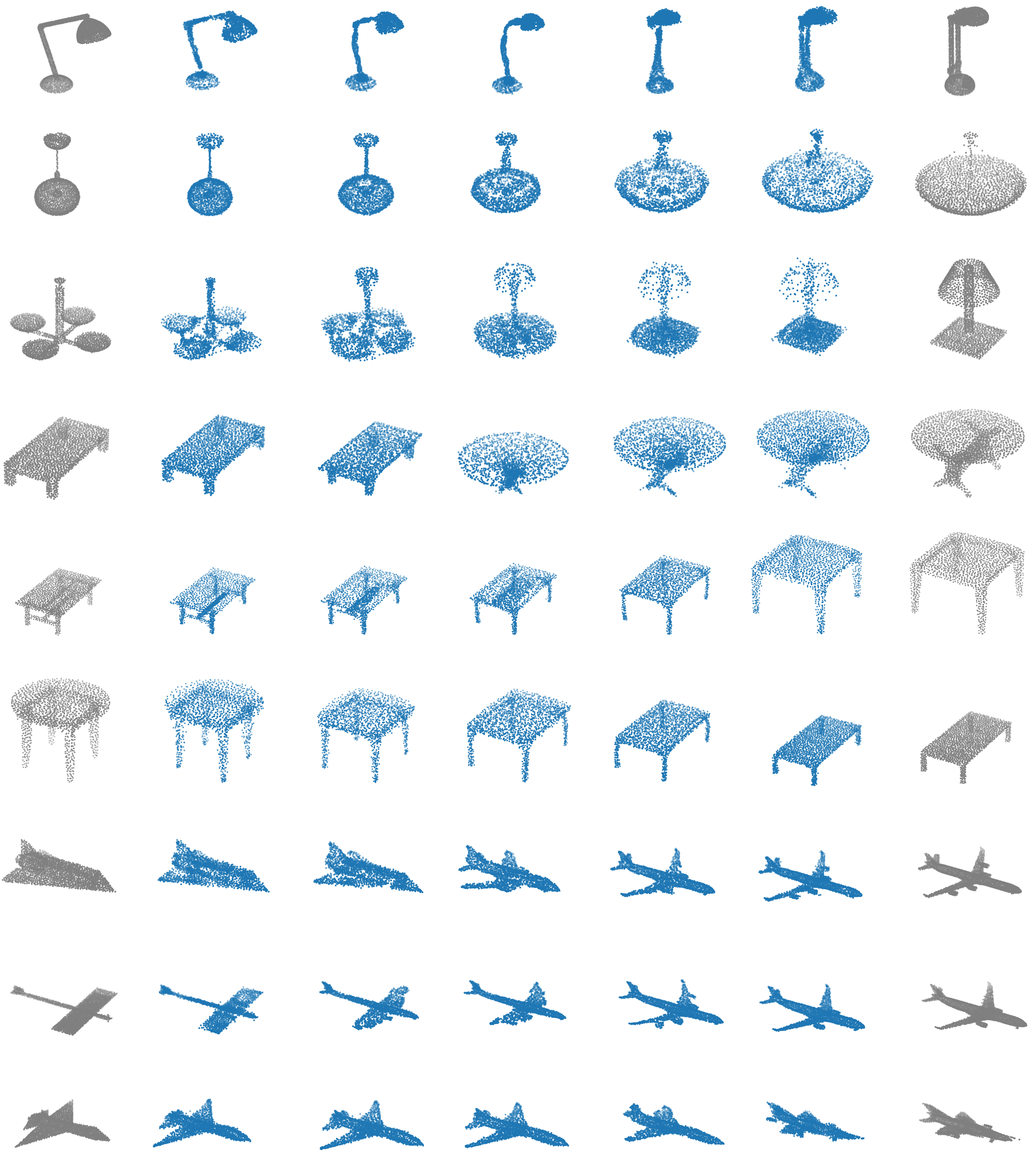}
\end{center}
  \caption{Additional qualitative results for shape interpolation (morphing) by ShapeInversion.  A sound transition from one shape to another is achieved by interpolation between their corresponding latent vectors $z$ and generator parameters $\theta$}
\label{fig:morph}
\end{figure*}

We show more qualitative results on shape completion for virtual scans (Fig.~\ref{fig:benchmark}), cross-data validation (Fig.~\ref{fig:cross_data2}), real scans (Fig.~\ref{fig:realscan2}), shape jittering (Fig.~\ref{fig:jit}), shape morpging (Fig.~\ref{fig:morph}), as well as on the effectiveness of PatchVariance on r-GAN (Fig.~\ref{fig:uniform_rgan}).

\newpage


\FloatBarrier
{\small
\bibliographystyle{ieee_fullname}
\bibliography{egbib}
}

\end{document}